\documentclass[10pt,letterpaper]{article}

\usepackage{cogsci}
\usepackage[dvipsnames]{xcolor}

\cogscifinalcopy 

\usepackage{pslatex}
\usepackage{apacite}
\usepackage{float} 

\usepackage{amsmath} 
\usepackage{bm} 
\usepackage{graphicx}  
\usepackage{booktabs}  

\usepackage{cleveref}  
\crefformat{section}{\S#2#1#3}
\crefformat{subsection}{\S#2#1#3}
\crefformat{subsubsection}{\S#2#1#3}
\crefname{figure}{Figure}{Figure}
\crefname{table}{Table}{Table}

\usepackage{xspace}

\usepackage{mwe}
\usepackage{subfig}

\title{Emergent Communication with Attention}

\author{
  {\large \bf Ryokan Ri (ryou0634@email.com)} \\
  {\large \bf Ryo Ueda (ryoryoueda@is.s.u-tokyo.ac.jp)} \\
  {\large \bf Jason Naradowsky (jason.narad@gmail.com)} \\
  The University of Tokyo, Japan
}

\begin{document}

\maketitle

\newcommand{\minisection}[1]{\noindent{\bf {#1}.}}
\newcommand{\minisectionNoDot}[1]{\vspace{2mm}\noindent{\bf {#1}}}

\newcommand{\Appendix}[1]{Appendix \ref{#1}}

\newcommand{\revised}{\color{red}}
\newcommand{\revisedend}{\color{black}}

\newcommand{\ie}{i.e.}
\newcommand{\eg}{e.g.}
\newcommand{\ia}{i.a.}

\newcommand{\matr}[1]{\bm{#1}}
\newcommand{\vect}[1]{\bm{#1}}
\newcommand{\KL}[2]{D_{\mathrm{KL}} \left( #1 \Vert #2 \right)}
\newcommand{\JS}[2]{D_{\mathrm{JS}} \left( #1 \Vert #2 \right)}
\newcommand{\Entropy}[1]{H \left( #1 \right)}
\newcommand{\Speaker}{Speaker\xspace}
\newcommand{\Listener}{Listener\xspace}

\newcommand{\TrainAcc}{TrainAcc\xspace}
\newcommand{\GenAcc}{GenAcc\xspace}
\newcommand{\TopSim}{TopSim\xspace}
\newcommand{\PosDis}{PosDis\xspace}
\newcommand{\BosDis}{BosDis\xspace}

\newcommand{\NoAttn}{NoAT\xspace}
\newcommand{\Attn}{AT\xspace}

\newcommand{\symb}[1]{\texttt{{#1}}}

\begin{abstract}
To develop computational agents that better communicate using their own emergent language, we endow the agents with an ability to focus their attention on particular concepts in the environment.
Humans often understand an object or scene as a composite of concepts and those concepts are further mapped onto words.
We implement this intuition as cross-modal attention mechanisms in \Speaker and \Listener agents in a referential game and show attention leads to more compositional and interpretable emergent language.
We also demonstrate how attention aids in understanding the learned communication protocol by investigating the attention weights associated with each message symbol and the alignment of attention weights between \Speaker and \Listener agents.
Overall, our results suggest that attention is a promising mechanism for developing more human-like emergent language.

\textbf{Keywords:}
emergent communication; attention; language compositionality
\end{abstract}

\maketitle

\section{Introduction}
Language is a defining characteristic of human beings, allowing us to efficiently convey a wide range of ideas.
One key aspect of language is compositionality, the ability to represent complex concepts through the combination of atomic units such as morphemes or words.
To understand the development of compositionality in human language, a field of research called \emph{emergent communication}~\cite{Lazaridou2020EmergentMC} studies the communication protocols developed by computational agents.
These agents are often constructed using artificial neural networks and optimized to solve a task requiring inter-agent communication.
During the optimization process, their ``language'' emerges.

Previous research has identified factors to potentially enhance the compositionality of emergent language, such as inter-generational learning~\shortcite{NEURIPS2019_b0cf188d,DBLP:conf/iclr/RenGLCK20} or introducing noise to the communication channel~\shortcite{NEURIPS2021_c2839bed}.
However, agent model architectures have typically had minimal assumptions about inductive bias, with speaker agents encoding information into a single vector to initialize a RNN decoder and generate symbols~\cite{DBLP:conf/iclr/LazaridouPB17,DBLP:conf/aaai/MordatchA18,DBLP:conf/iclr/RenGLCK20}.
While some studies have explored architectural variations~\shortcite{Sowik2020StructuralIB,DBLP:conf/iclr/EvtimovaDKC18}, there is still much to be learned about the effects of different architectures on inductive bias, particularly those that reflect human cognitive processes.

In this study, we explore \emph{the attention mechanism}.
The conceptual core of attention is the ability to adaptively focus on relevant information, and attention has been shown to play an important role in human cognition~\cite{Rensink2000TheDR}, language development~\cite{deDiegoBalaguer2016TemporalAA}, and intentional communication~\cite{Brinck2000AttentionAT}.
Introducing the notion of attention into emergent communication can enhance its resemblance to human communication and expand the scope of the research field.
In this paper, we hypothesize that attention can help agents learn clear associations between subparts of input stimuli and language symbols, resulting in more compositional language.

Another reason to explore the attention mechanism is its interpretability.
Emergent language is usually optimized to maximize task rewards and the learned communication protocol often results in counter-intuitive and opaque encoding~\cite{bouchacourt-baroni-2018-agents}.
Several metrics have been proposed to measure specific characteristics of emergent language~\cite{Brighton2006UnderstandingLE,Lowe2019OnTP} but these metrics provide rather a holistic view of emergent language and do not tell us a fine-grained view of what each symbol is meant for or understood as.
Attention weights, on the other hand, have been shown to provide insights into the basis of the network's prediction~\shortcite{DBLP:journals/corr/BahdanauCB14,DBLP:conf/icml/XuBKCCSZB15,Yang2016StackedAN}.
Incorporating attention in the process of symbol production/comprehension will allow us to inspect the meaning of each symbol in the messages.

In this paper, we test attention agents with the visual referential game~\cite{Lewis1969-LEWCAP-4,DBLP:conf/iclr/LazaridouPB17}, which involves two agents: {\it \Speaker} and {\it \Listener}.
The goal of the game is to convey the types of items in an image that \Speaker sees to \Listener.
To offer extensive empirical results, we experiment with two types of popular network architectures, LSTM~\cite{HochSchm97} and Transformer~\cite{DBLP:conf/nips/VaswaniSPUJGKP17}, to implement the agents.
We compare the attention agents against their non-attention counterparts to show that adding attention mechanisms to either/both \Speaker or/and \Listener helps develop a more compositional language.
We also examine the attention weights and investigate how they shed light on the learned language.

\begin{figure*}[t]
\centering
\includegraphics[width=14.0cm]{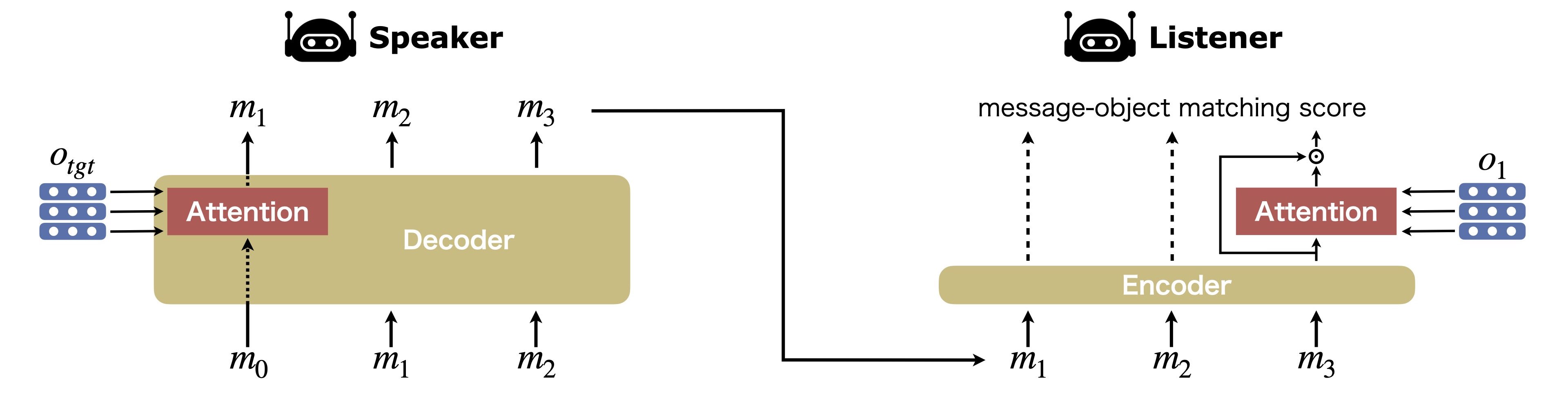}
\caption{Illustration of the attention agents in the referential game.}
\label{fig:attention-agents}
\end{figure*}

\section{Experimental Framework}
\label{sec:experimental_framework}

\subsection{Referential Game}
\label{subsec:referential_game}

We study emergent language in the referential game~\cite{Lewis1969-LEWCAP-4,DBLP:conf/iclr/LazaridouPB17}.
The game focuses on a basic feature of language, referring to things.
The version of the referential game used in this paper is structured as follows:

\begin{enumerate}
  \item \Speaker is presented with a target object $o_{tgt} \in \mathcal{O}$ and generates a message $m$ that consists of a sequence of symbols.
  \item \Listener receives the message $m$ and a candidate set $C = \{o_{1}, o_{2}, ..., o_{|C|}\}$ including the target object $o_{tgt}$ and distractor objects sampled randomly without replacement from $\mathcal{O}$.
  \item \Listener chooses one object from the candidate set and if it is the target object, the game is considered successful.
\end{enumerate}

The objects can be represented as a set of attributes.
The focus here is whether agents can represent the objects in a compositional message based on the attributes.

\subsection{Agent Architectures}
Our goal in this paper is to test the effect of the attention mechanism on emergent language.
The attention mechanism in machine learning takes a query vector~$\vect{x}$ and key-value vectors~$\{\vect{y}_1, ..., \vect{y}_L\}$ as input.
The key-value vectors are optionally transformed into separate keys and values.
Attention scores~$\{s_1, ..., s_L\}$ are calculated as the similarity between the query and keys to produce the attention weights via the softmax function.
Finally, the attention weights are used to produce a weighted sum of the value vectors.

A key feature of attention is that it allows the agents to selectively attend to a part of multi-vector inputs.
Our hypothesis is that by modeling direct associations between symbols as queries and multi-vector inputs as key-value, agents will tend to assign symbols to meaningful subparts of the inputs instead of opaque and non-compositional information.
To test this hypothesis, we design non-attention and attention agents for both \Speaker and \Listener (\cref{fig:attention-agents}).

\subsubsection{Object Encoder}
The objects are presented to the agents in the form of a set of real-valued vectors, in our case, patch-wise pretrained CNN feature vectors, as $\{\vect{o}^1, ..., \vect{o}^A\}$.

\Speaker and \Listener agents have their individual object encoders.
The input vectors are independently linear-transformed into the size of the agent's hidden size and then successively go through the \texttt{gelu} activation~\cite{Hendrycks2016GaussianEL}, which is commonly adapted in Transformer-based neural networks.
The non-attention agents average the transformed vectors into a single vector ${\hat{\vect{o}}}$ for the subsequent computations, whereas the attention agents leave the vectors intact and will attend to the set of vectors $\{\hat{\vect{o}}^1, ..., \hat{\vect{o}}^A\}$.

\subsubsection{Speaker Agents}
\Speaker has a message decoder that takes the encoded vector(s) of the target object as input and generates a multi-symbol message~$m = (m_1, ..., m_T)$.
To provide extensive empirical evidence on the effect of the attention mechanism, we experiment with two common decoder architectures: the {\bf LSTM} decoder from \citeA{luong-etal-2015-effective} and the {\bf Transformer} decoder from \citeA{DBLP:conf/nips/VaswaniSPUJGKP17}.

At each time step $t$, the decoders embed a previously generated symbol into a vector $\vect{m}_{t-1}$ and produce an output hidden vector through three steps: (1) contextualization; (2) attention; (3) post-processing.
As the decoders basically follow the original architecture, we only briefly describe each step in the LSTM and Transformer decoder with emphasis on how attention is incorporated.
The contextualization step updates the input vector with the information of previous inputs.
The LSTM decoder uses a LSTM cell~\cite{HochSchm97} and the Transformer decoder uses the self-attention mechanism~\cite{DBLP:conf/nips/VaswaniSPUJGKP17}.

Then, with the contextualized input vector as the query~$\vect{x}_t$ and the object vectors as the key-value vectors~$\{\hat{\vect{o}}^1, ..., \hat{\vect{o}}^A\}$, the decoders perform attention.
The LSTM decoder uses the bilinear attention, where the attention score $s_i$ is computed as $s_t^i = \vect{x}_t^\top \matr{W}_b \hat{\vect{o}}^i$, where $\matr{W}_b$ is a learnable matrix.
The attention vector is calculated as the weighted sum of the original key-value vectors.
The Transformer attention first linear-transforms the input vector as $\vect{q}_t = \matr{W}_q \vect{x}_t, \vect{k}^i = \matr{W}_k \hat{\vect{o}}^i, \vect{v}^i = \matr{W}_v \hat{\vect{o}}^i$, where $\matr{W}_q, \matr{W}_k, \matr{W}_v$ are learnable matrices.
Then the attention score is calculated using the scaled dot attention: $s_t^i = (\vect{q}_t^\top \vect{k}^{i}) / \sqrt{d}$, where $d$ is the dimension of the query and key vectors.
Finally, the attention vector is calculated as the weighted sum of the value vectors $\vect{v}^i$.
The original Transformer also has the multi-head attention mechanism, but in the main experiments, we set the number of the attention heads to one for interpretability and ease of analysis.

In the post-processing step, the original query vector~$\vect{x}_t$ and attended vector $\hat{\vect{x}}_t$ combine to generate the hidden vector to predict the next symbol.
This integration often includes vector concatenation, addition, and transformation, as detailed in the original papers~\cite{luong-etal-2015-effective, DBLP:conf/nips/VaswaniSPUJGKP17}.

\minisectionNoDot{Non-attention (\NoAttn) \Speaker} is a baseline agent that encodes the target object as a single vector $\hat{\vect{o}}_{tgt}$.
In the decoder's source-target attention, this vector is always attended to, and the focus remains fixed throughout message generation.

\minisectionNoDot{Attention (\Attn) \Speaker}, in contrast, encodes the target object into a set of vectors $\{\hat{\vect{o}}_{tgt}^1, ..., \hat{\vect{o}}_{tgt}^A\}$ and the source-target attention dynamically changes its focus at each time step.

Our agent design aims to provide a fair comparison between non-attention and attention agents, by ensuring they possess the same modules and number of parameters.
The only distinction is the latter's ability to adjust its focus dynamically when generating each symbol.

\subsubsection{Listener Agents}
\Listener tries to predict the target object from a set of candidate objects $C = \{o_{1}, o_{2}, ..., o_{|C|}\}$ given the speaker message $m$ by computing message-object matching scores $\{s_1, ..., s_{|C|}\}$ and choosing the object with the maximum score.
\Listener first encodes the objects using the object encoder and also encodes each symbol in the message into vectors $\{\vect{m}^1, ..., \vect{m}^{T}\}$ using a message encoder, for which the LSTM-based agent uses the bidirectional LSTM and the Transformer-based agent uses the Transformer encoder.

\minisectionNoDot{Non-attention (\NoAttn) \Listener} encodes each candidate object into a single vector ${\hat{\vect{o}}}_i$.
The agent also averages the encoded symbol vectors into a single vector $\vect{m} = \frac{1}{T} \sum_{i=1}^T \vect{m}^i$.
The message-object matching score is computed by taking the dot product of the object and message vector $s_i = \hat{\vect{o}}_i^\top \vect{m}$.

\minisectionNoDot{Attention (\Attn) \Listener} encodes each object into a set of attribute vectors $\{\hat{\vect{o}}_i^1, ..., \hat{\vect{o}}_i^A\}$ and use the encoded symbol vectors as it is.
With each encoded symbol vector $\vect{m}^t$ as query, the model produces an attention vector $\hat{\vect{m}}^t_i$ with the object attribute vectors as key-value using the dot-product attention.
Intuitively, the attention vector $\hat{\vect{m}}^t_i$ is supposed to represent the attributes of the object $o_i$ relevant to the symbol $m^t$.
Then, the symbol-object matching scores are computed by taking the dot product between the attention vector and each symbol vector: $s_i^t = \hat{\vect{m}}_i^{t\top} \vect{m}^t$.
Finally, the symbol-object matching scores are averaged to produce the message-object matching score: $s^i = \frac{1}{T} \sum_t s_i^t$.

\subsection{Optimization}
The parameters of \Speaker~$\theta_{S}$ and \Listener~$\theta_{L}$ are both optimized toward the task success.

\Speaker is trained with the REINFORCE algorithm~\cite{REINFORCE1992}.
The message decoder produces the probability distribution of which symbol to generate $\pi_{\theta_{S}}(\cdot|t)$ at each time step ${t}$.
At training time, message symbols are randomly sampled according to the predicted probabilities and the loss function for the \Speaker message policy is $\mathcal{L}_{\pi}(\theta_{S}) = \sum_{t} r \log(\pi_{\theta_{S}}(m_t| t))$ where $m_t$ denotes the $t$-th symbol in the message.
The reward $r$ is set to 1 if \Listener selects the correct target object from the candidate set and 0 otherwise.

As an auxiliary loss function, we employ an entropy regularization loss ${L}_{H}(\theta_{S}) = - \sum_{t} \Entropy{\pi_{\theta_{S}}(\cdot|t)}$, where $H$ is the entropy of a probability distribution, to encourage exploration.
We also add a KL loss ${L}_{\text{KL}}(\theta_{S}) = \sum_{t} \KL{\pi_{\theta_{S}}(\cdot| t)}{\pi_{\bar{\theta}_{S}}(\cdot| t)}$, where the policy $\pi_{\bar{\theta}_{S}}$ is obtained by taking an exponential moving average of the weights of $\theta_{S}$ over training, to stabilize the training~\cite{DBLP:conf/iclr/ChaabouniSATTDM22}.
In summary, the final speaker loss is $\mathcal{L}({\theta_{S}}) = \mathcal{L}_{\pi}({\theta_{S}}) + \alpha \mathcal{L}_{H}({\theta_{S}}) + \beta \mathcal{L}_{KL}({\theta_{S}})$, where $\alpha$ and $\beta$ are hyperparameters.

\Listener is trained with a multi-class classification loss.
The message-object matching scores are converted through the softmax operation to $p_{\theta_{L}}(o_i | C)$, the probability of choosing the object $o_i$ as the target from the candidate set $C$.
Then \Listener is trained to maximize the probability of predicting the target object by minimizing the loss function $\mathcal{L}(\theta_{L}) = - \log{p_{\theta_{L}}(o_{tgt} | C)}$.

\subsection{Evaluation Metrics}
We quantitatively evaluate emergent languages from how well the language can be used to solve the task and how well the language exhibits compositionality.

\minisectionNoDot{Training accuracy (\TrainAcc)} measures the task performance with objects seen during training.
This indicates how the agent architectures are simply effective to solve the referential game.

\minisectionNoDot{Generalization accuracy (\GenAcc)} measures the task performance with objects unseen during training.
We split the distinct object types in the game into train and evaluation sets and the generalization accuracy is computed with the evaluation set.
As each object can be represented as a combination of attribute values, what we expect for the agents is to learn to combine symbols denoting each attribute value in a systematic way so that the language can express unseen combinations of known attribute values.

\newcommand{\figWidth}{4.8cm}

\begin{figure*}[t]
\begin{minipage}{.33\linewidth}
\centering
\subfloat[\TrainAcc]{\label{fig:main-fashion-train}\includegraphics[width=\figWidth]{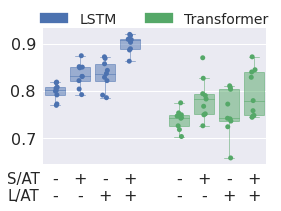}}
\end{minipage}\hfill%
\begin{minipage}{.33\linewidth}
\centering
\subfloat[\GenAcc]{\label{fig:main-fashion-gen}\includegraphics[width=\figWidth]{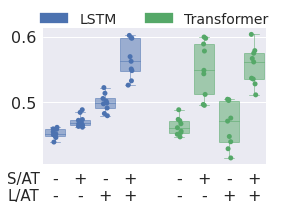}}
\end{minipage}\hfill%
\begin{minipage}{.33\linewidth}
\centering
\subfloat[\TopSim]{\label{fig:main-fashion-topsim}\includegraphics[width=\figWidth]{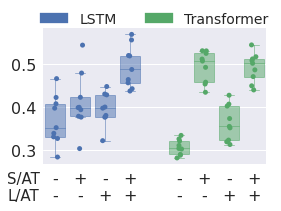}}
\end{minipage}\par\medskip 
\vspace{-3mm}
\caption{The results of the referential game. The color of the boxes indicates the base architecture of the agents (LSTM or Transformer) and the x-axis labels indicates whether \Speaker and \Listener use attention.}
\label{fig:main-results}
\vspace{-3mm}
\end{figure*}

\minisectionNoDot{Topographic similarity (\TopSim)}, also known as Representational Similarity Analysis~\cite{Kriegeskorte2008}, is one of the most commonly used metrics to assess the compositionality of emergent language~\cite{Brighton2006UnderstandingLE,DBLP:conf/iclr/LazaridouHTC18,DBLP:conf/iclr/RenGLCK20,chaabouni-etal-2020-compositionality}.
Intuitively, \TopSim checks if similar objects have similar messages assigned.
To compute \TopSim, we enumerate all the object-message pairs $\{(o^1, m^1), ...., (o^{|\mathcal{O}|}, m^{|\mathcal{O}|})\}$ with a trained \Speaker and define a distance function for objects $d_{\mathcal{O}}(o^i, o^j)$ and messages $d_{\mathcal{M}}(m^i, m^j)$.
Then we compute Spearman's correlation between pairwise distances in the object and message space.
For the distance function for objects $d_{\mathcal{O}}(o^i, o^j)$, we use the cosine distance of the binary attribute value vectors and for the distance function of messages $d_{\mathcal{M}}(m^i, m^j)$ the edit distance of message symbols.

\section{Experimental Setup}
\label{sec:exp-setup}

\subsection{Fashion-MNIST Game}
\label{subsec:fashion-mnist}

Attention has been shown to be able to associate a symbol and a relevant region of an image to solve the task~\cite{DBLP:conf/icml/XuBKCCSZB15,Yang2016StackedAN}.
To develop human-like emergent languages, one important question is whether the attention agents develop an interpretable language such that we can understand the meaning of each symbol by inspecting the attended region.
For this purpose, we design a multi-item image referential game using the Fashion-MNIST dataset\footnote{https://github.com/zalandoresearch/fashion-mnist}~\cite{Xiao2017FashionMNISTAN}.

The game objective is to communicate a type of object between \Speaker and \Listener.
Each object is defined as a combination of two classes from the Fashion-MNIST dataset (\eg, {\it T-shirt} and {\it Sneaker}).
As the dataset has 10 classes, there are a total of $_{10} \mathrm{C}_2$ = 45 object types.
These 45 types are randomly split into training and evaluation sets at a ratio of 30/15, and the number of candidates is set to 15.

The FashionMNIST items on the images are spatially disentangled and ideal for evaluating attention mechanisms.
While learning to attend to abstract attributes such as color and shape in images can be more challenging, we anticipate that it is achievable with the use of high-quality feature extractors and learning configurations.

\subsection{Input Representation}
\begin{figure*}[t]
\begin{minipage}{.33\linewidth}
\centering
\subfloat[Successful coordination with the same image.]{\label{fig:attention-image:a}\includegraphics[height=2.6cm]{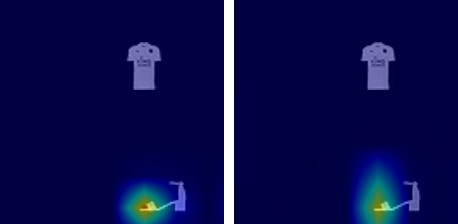}}
\end{minipage}\hfill%
\begin{minipage}{.33\linewidth}
\centering
\subfloat[Successful coordination with the different images with the same type.]{\label{fig:attention-image:c}\includegraphics[height=2.6cm]{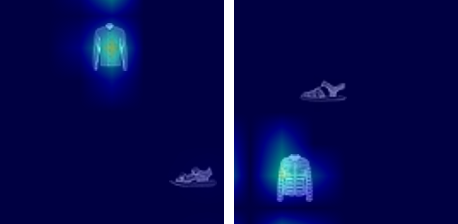}}
\end{minipage}\hfill%
\begin{minipage}{.33\linewidth}
\centering
\subfloat[Failure case. The attention maps are not focused.]{\label{fig:attention-image:e}\includegraphics[height=2.6cm]{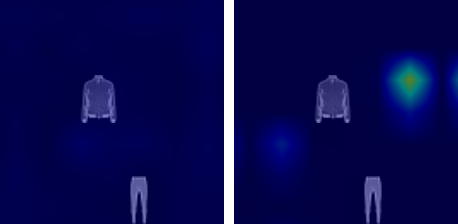}}
\end{minipage}\hfill%
\caption{Attention maps produced by a \Attn-\Attn agent pair (Left: \Speaker, Right: \Listener).}
\label{fig:attention-images}
\vspace{-3mm}
\end{figure*}

Each object is presented to the agents as feature vectors extracted from a pixel image.
The image is created by placing on a $224\times224$ black canvas two item images, each of which is rescaled to the size of $48\times48$ (\cref{fig:attention-images}).
The places are randomly sampled so that the items never overlap.

The specific item images and their positions are randomly sampled every time the agents process the objects both during training and evaluation time to avoid degenerated solutions that exploit spurious features of an image~\cite{DBLP:conf/iclr/LazaridouHTC18,bouchacourt-baroni-2018-agents}.
Also, the \Speaker and \Listener are presented as the target object with two images depicting the same item types, but with different instances and locations, to facilitate learning a robust communication protocol~\shortcite{rodriguez-luna-etal-2020-internal,DBLP:conf/nips/MuG21}.

Each image is encoded into $7\times7\times768$-dim feature vectors with a pretrained ConvNet\footnote{The pretrained model is registered as \texttt{convnext\_tiny} in the \texttt{torchvision} library}~\cite{liu2022convnet}.
For non-attention models, the feature vectors are averaged across spatial axes into a single 768-dim feature vector.

\subsection{Agent Configurations}
The vocabulary size of the agents is set to 20 and the message length is 2.
A perfectly compositional language would refer to each item in the image with different symbols with a consistent one-to-one mapping.
The sizes of embeddings and hidden sizes of the \Speaker and \Listener are set to 256.

The hyperparameters (\cref{tb:configs}) are tuned for the entropy loss weight $\alpha$.
The reported scores are obtained from the top 10 agents in terms of the generalization score for each setting.

\begin{table}[ht]
    \centering
    \begin{tabular}{lc}
        \toprule
        Training Batch size                   & 480   \\
        Max Training steps                    & 50K   \\
        Evaluation Rounds                     & 15000 \\
        Entropy loss weight $\alpha$          & [0.1, 0.01, 0.001] \\
        KL loss weight $\beta$                & 0.1   \\
        Learning rate                         & 1e-4  \\
        \bottomrule
    \end{tabular}
    \caption{Hyperparameters of the experiments.}
    \label{tb:configs}
\end{table}

\begin{figure}[t]
\begin{minipage}{.5\linewidth}
\centering
\subfloat[Speaker]{\label{fig:symbol-mapping:c}\includegraphics[height=3.4cm]{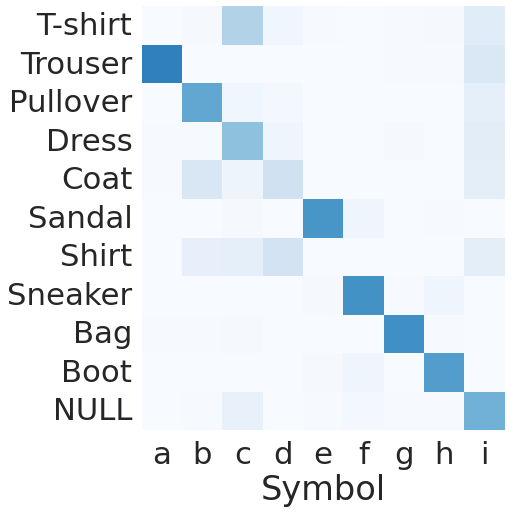}}
\end{minipage}\hfill%
\begin{minipage}{.5\linewidth}
\centering
\subfloat[Listener]{\label{fig:symbol-mapping:d}\includegraphics[height=3.4cm]{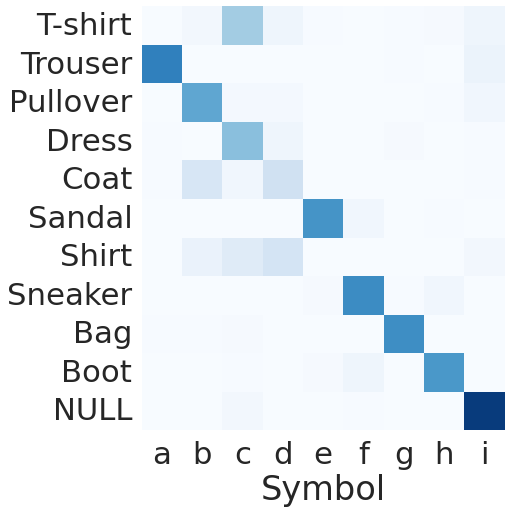}}
\end{minipage}\hfill%
\caption{A frequency heatmap of symbol-concept association derived from the attention weights of Transformer \Attn-\Attn agent pairs.}
\label{fig:attention-mapping}
\vspace{-3mm}
\end{figure}

\section{Results}
\subsection{Attention agents find more compositional solutions}
\label{subsec:result-main}

We evaluate non-attention agents and attention agents where either/both \Speaker and \Listener have dynamic attention (\cref{fig:main-results}).
We will not discuss the distinctions between LSTM and Transformer as they vary in multiple aspects of their architecture.
We focus on the difference between the non-attention and attention agents within each architecture.

We observe a general trend that the attention agents perform better than the non-attention baseline (\NoAttn-\NoAttn), which is indicated by the better average scores in task generalization (\GenAcc) and compositionality metrics (\TopSim).
This observation provides evidence in support of the hypothesis that the attention mechanism creates pressure for learning more compositional emergent languages.

One possible interpretation is that the attention mechanism adds flexibility to the model and it simply leads to better learning of the task.
We observe the effect in \TrainAcc, where the attention models consistently outperform their non-attention baselines.
However, we can still see the contribution of the attention mechanism besides the flexibility.
We can see some \NoAttn-\NoAttn agents and \Attn-\Attn agents exhibit comparable \TrainAcc around 75\%, which means they are successful at optimization to a similar degree.
However, we observe all the \Attn-\Attn agents significantly outperform any of the \NoAttn-\NoAttn agents, which indicates the degree of optimization alone cannot explain the better \GenAcc and \TopSim scores of the attention model.
Therefore, the results supports our initial hypothesis that the attention mechanism facilitates developing compositional languages.

In some settings, we observe that adding the attention mechanism to only one of Speaker and Listener does not lead to better \TopSim scores compared the \NoAttn-\NoAttn agents to as in the \Attn-\NoAttn and \NoAttn-\Attn LSTM agents.
However, when both Speaker and Listener agents have the attention mechanism, they all exhibit more generalizable and compositional languages, which indicates the effect of attention in Speaker and Listener is multiplicative.

\subsection{Attention agents learn to associate input attributes and symbols}
Having confirmed that the attention agents give rise to more compositional languages, we proceed to examine if they use attention in an expected way, \ie, producing/understanding each symbol by associating them with a single input concept.
We focus on analyzing the Transformer agents below.

We inspect the attention weights and confirm that attention agents generally learn to focus on a single object when generating a symbol as in \cref{fig:attention-images}(a) and (b), although there are some failure cases as shown in \cref{fig:attention-images}(c).

Given the observations above, we can associate each symbol in each message with the concepts defined in the game via attention weights.
We visualize the association from a pair of \Attn-\Attn agents in \cref{fig:attention-mapping} to inspect the mapping patterns developed by the agents.
A symbol is considered to be associated with a concept when the center of gravity of the attention weights is within the bounding box of the item in an image.
The results show that the mappings learned by \Speaker and \Listener have a strong tendency to agree in almost all cases.
We identify three types of symbol-to-concept mapping patterns.

\minisection{Monosemy}
A single symbol always refers to a single concept, \eg, \symb{a}, \symb{e}, and \symb{g}.
This is a desired mapping pattern that allows an unambiguous interpretation of symbols.

\minisection{Polysemy}
A single symbol refers to multiple concepts, \eg, \symb{b}, \symb{c}, and \symb{d}.
These symbols are somewhat ambiguous, but they seem to be affected by the visual similarity of the fashion items, \eg, \symb{b} refers to tops (Pullover, Coat, and Shirt) and \symb{f} refers to shoes (Sandal, Sneaker, and Boot).
This pattern demonstrates that the semantics of emergent language can be heavily influenced by the property of the input objects.

\minisection{Gibberish}
We observe a few cases where the attention weights do not consistently focus on any particular regions, \eg, the symbol \symb{i}.
These gibberish symbols could have conveyed something informative but uninterpretable to humans, but we confirmed that this indicates the results of optimization failure.
We run 10 rounds of referential games with 45 candidates with a single agent pair, and the communication success rate is much lower when the Speaker's message contained gibberish symbols, compared to the overall score ($27.7 < 45.3$).

\subsection{The coordination of \Speaker and \Listener attentions predicts the task performance}
\label{subsec:attention-discrepancy}
An important prerequisite of successful communication is the participants engaging in joint attention and establishing a shared understanding of each word~\cite{garrod_2004}.
Here we show that the degree of the alignment between \Speaker's and \Listener's attention weights can be regarded as a proxy of their mutual understanding and predictive of communication success.

We define a metric called {\it attention discrepancy}, which measures the difference between the attention weights of \Speaker and \Listener given the same inputs.
For each symbol $m_t$ in a message, \Speaker and \Listener have attention weights $\vect{a}^{(S)}_{t}, \vect{a}^{(L)}_{t}$.
The metric is calculated by computing the average of the Jensen–Shannon Divergence~\cite{Lin1991DivergenceMB} between the attention maps: $\frac{1}{T} \sum_{t=1}^{T} \JS{\vect{a}^{(S)}_{t}}{\vect{a}^{(L)}_{t}}$.

\begin{figure}[t]
\centering
\includegraphics[height=3.2cm]{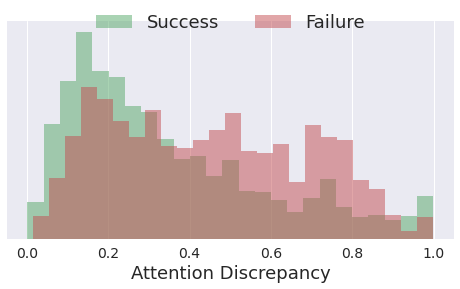}
\caption{The distribution of attention discrepancy scores from the trials of the \Attn-\Attn agents. The frequency is normalized within successful and failed trials respectively.}
\label{fig:attention-discrepancy}
\end{figure}

\cref{fig:attention-discrepancy} shows the distributions of attention discrepancy scores within successful and failed communications.
These distributions are significantly dissimilar ($p < 0.001$ in the Kolmogorov-Smirnov test), with successful communications exhibiting lower scores than failed communications
The results indicate a correlation between attention weight alignment and communication success to some degree. This suggests that attention agents develop intuitive communication protocols that rely on a shared understanding of symbols.
\section{Related Work}

\subsection{Emergent Language}
Emergent language should exhibit {\it compositionality}, allowing interpretability, generalizability, and ease of learning~\cite{NEURIPS2019_b0cf188d,DBLP:conf/iclr/RenGLCK20}.
Existing studies have shown that the compositionality of language can be improved by learning across generations~\shortcite{NEURIPS2019_b0cf188d,DBLP:conf/iclr/RenGLCK20}, learning with a population~\shortcite{DBLP:conf/iclr/RitaSGPD22}, applying noise to the communication channel~\cite{NEURIPS2021_c2839bed}, and balancing the learning speed of the agents~\cite{Rita2022EmergentCG}.

While recent studies have primarily used simple RNN-based \Speaker and \Listener agents, other agent architectures have been employed for specific purposes.
For example, \shortciteA{chaabouni-etal-2019-word} and \shortciteA{ueda-etal-2022-cgi} used the LSTM sequence-to-sequence (with attention) architecture to study transduction from a grammar-generated input to an emergent language.
\citeA{DBLP:conf/iclr/EvtimovaDKC18} compared attentional and non-attentional agents in a multi-modal and multi-step referential game, observing improvements in the out-of-domain test but not the in-domain test.
However, it is unclear whether these trends generalize to other settings.

Our study aims to investigate the impact of the model architecture's inductive biases on the compositionality of the emergent language.
Similarly, \citeA{Sowik2020StructuralIB} compared \Speaker agents that process inputs as a graph, sequence, or bag-of-words to show that the graph architecture results in more compositional emergent languages.
Our study contributes further empirical evidence in this direction, specifically exploring the role of the attention mechanism.

\subsection{Attention Mechanism in Machine Learning}
The attention mechanism has proven effective in supervised learning~\cite{DBLP:conf/icml/XuBKCCSZB15,DBLP:conf/nips/VaswaniSPUJGKP17}, becoming an integral part of modern neural networks.
Conceptually, this mechanism models pairwise associations between a query and a subset of key-values.
This enables the model to focus on a subpart of compositional representation and has been shown to enforce compositional solutions in visual reasoning~\cite{DBLP:conf/iclr/HudsonM18}, symbolic reasoning~\cite{korrel-etal-2019-transcoding}, image generation~\cite{Hudson2021CompositionalTF}.
Our study shows that by utilizing attention to model the association between a symbol and an object attribute, the agents can discover more compositional languages in a referential game without additional supervision.

Attention may also offer interpretability.
Attention has been shown to provide plausible alignment patterns between inputs and outputs, such as source and target words in machine translation~\cite{DBLP:journals/corr/BahdanauCB14} and image regions and words in image captioning~\cite{DBLP:conf/icml/XuBKCCSZB15,Yang2016StackedAN}.
However, the alignment patterns may not always align with human intuitions~\cite{alkhouli-etal-2018-alignment,Liu2020ProphetAP} and there is an ongoing debate about the extent to which the attention weights can be used as explanations for model predictions~\cite{jain-wallace-2019-attention,wiegreffe-pinter-2019-attention,bibal-etal-2022-attention}.
In this paper, our agents exhibit relatively straightforward attention patterns that allow for unambiguous interpretations.
However, the interpretability of attention weights with more complex models and environments requires further investigation.

\section{Discussion and Conclusion}
When we seek to interpret a complex object, we are likely to dynamically change our focus on its subpart to describe the whole~\cite{Rensink2000TheDR}.
Hearing a word aids in recognizing its referent \cite{Boutonnet2015WordsJV} and affects where to focus in an image~\cite{Estes2008HeadUF}.
Motivated by these observations, we implemented agents with the dynamic interaction between symbols and inputs in the form of the attention mechanism.
We showed that the attention agents develop more compositional languages than their non-attention counterparts.
This implies that the human capacity for dynamic focus may have contributed to developing compositional language.

To better understand how cognitive properties shape language, future research should explore additional architectural variations inspired by human cognitive processing.
For instance, incorporating joint attention~\shortcite{Kwisthout2008JointAA} into language training may enhance communication success, as alignment between speaker and listener attention appears to be a factor.

\section*{Acknowledgement}
We thank the anonymous reviewers for their insightful and helpful comments to improve the manuscript.

\bibliographystyle{apacite}

\setlength{\bibleftmargin}{.125in}
\setlength{\bibindent}{-\bibleftmargin}

\bibliography{references}

\begin{thebibliography}{}

\bibitem [\protect \citeauthoryear {%
Alkhouli%
, Bretschner%
\BCBL {}\ \BBA {} Ney%
}{%
Alkhouli%
\ \protect \BOthers {.}}{%
{\protect \APACyear {2018}}%
}]{%
alkhouli-etal-2018-alignment}
\APACinsertmetastar {%
alkhouli-etal-2018-alignment}%
\begin{APACrefauthors}%
Alkhouli, T.%
, Bretschner, G.%
\BCBL {}\ \BBA {} Ney, H.%
\end{APACrefauthors}%
\unskip\
\newblock
\APACrefYearMonthDay{2018}{}{}.
\newblock
{\BBOQ}\APACrefatitle {On The Alignment Problem In Multi-Head Attention-Based
  Neural Machine Translation} {On the alignment problem in multi-head
  attention-based neural machine translation}.{\BBCQ}
\newblock
\BIn{} \APACrefbtitle {Proceedings of the Third Conference on Machine
  Translation: Research Papers.} {Proceedings of the third conference on
  machine translation: Research papers.}
\PrintBackRefs{\CurrentBib}

\bibitem [\protect \citeauthoryear {%
Bahdanau%
, Cho%
\BCBL {}\ \BBA {} Bengio%
}{%
Bahdanau%
\ \protect \BOthers {.}}{%
{\protect \APACyear {2015}}%
}]{%
DBLP:journals/corr/BahdanauCB14}
\APACinsertmetastar {%
DBLP:journals/corr/BahdanauCB14}%
\begin{APACrefauthors}%
Bahdanau, D.%
, Cho, K.%
\BCBL {}\ \BBA {} Bengio, Y.%
\end{APACrefauthors}%
\unskip\
\newblock
\APACrefYearMonthDay{2015}{}{}.
\newblock
{\BBOQ}\APACrefatitle {Neural Machine Translation by Jointly Learning to Align
  and Translate} {Neural machine translation by jointly learning to align and
  translate}.{\BBCQ}
\newblock
\BIn{} \APACrefbtitle {International Conference on Learning Representations.}
  {International conference on learning representations.}
\PrintBackRefs{\CurrentBib}

\bibitem [\protect \citeauthoryear {%
Bibal%
\ \protect \BOthers {.}}{%
Bibal%
\ \protect \BOthers {.}}{%
{\protect \APACyear {2022}}%
}]{%
bibal-etal-2022-attention}
\APACinsertmetastar {%
bibal-etal-2022-attention}%
\begin{APACrefauthors}%
Bibal, A.%
, Cardon, R.%
, Alfter, D.%
, Wilkens, R.%
, Wang, X.%
, Fran{\c{c}}ois, T.%
\BCBL {}\ \BBA {} Watrin, P.%
\end{APACrefauthors}%
\unskip\
\newblock
\APACrefYearMonthDay{2022}{}{}.
\newblock
{\BBOQ}\APACrefatitle {Is Attention Explanation? An Introduction to the Debate}
  {Is attention explanation? an introduction to the debate}.{\BBCQ}
\newblock
\BIn{} \APACrefbtitle {Proceedings of the 60th Annual Meeting of the
  Association for Computational Linguistics (Volume 1: Long Papers).}
  {Proceedings of the 60th annual meeting of the association for computational
  linguistics (volume 1: Long papers).}
\PrintBackRefs{\CurrentBib}

\bibitem [\protect \citeauthoryear {%
Bouchacourt%
\ \BBA {} Baroni%
}{%
Bouchacourt%
\ \BBA {} Baroni%
}{%
{\protect \APACyear {2018}}%
}]{%
bouchacourt-baroni-2018-agents}
\APACinsertmetastar {%
bouchacourt-baroni-2018-agents}%
\begin{APACrefauthors}%
Bouchacourt, D.%
\BCBT {}\ \BBA {} Baroni, M.%
\end{APACrefauthors}%
\unskip\
\newblock
\APACrefYearMonthDay{2018}{}{}.
\newblock
{\BBOQ}\APACrefatitle {How agents see things: On visual representations in an
  emergent language game} {How agents see things: On visual representations in
  an emergent language game}.{\BBCQ}
\newblock
\BIn{} \APACrefbtitle {Proceedings of the 2018 Conference on Empirical Methods
  in Natural Language Processing.} {Proceedings of the 2018 conference on
  empirical methods in natural language processing.}
\PrintBackRefs{\CurrentBib}

\bibitem [\protect \citeauthoryear {%
Boutonnet%
\ \BBA {} Lupyan%
}{%
Boutonnet%
\ \BBA {} Lupyan%
}{%
{\protect \APACyear {2015}}%
}]{%
Boutonnet2015WordsJV}
\APACinsertmetastar {%
Boutonnet2015WordsJV}%
\begin{APACrefauthors}%
Boutonnet, B.%
\BCBT {}\ \BBA {} Lupyan, G.%
\end{APACrefauthors}%
\unskip\
\newblock
\APACrefYearMonthDay{2015}{}{}.
\newblock
{\BBOQ}\APACrefatitle {Words jump-start vision: a label advantage in object
  recognition} {Words jump-start vision: a label advantage in object
  recognition}.{\BBCQ}
\newblock
\APACjournalVolNumPages{Journal of vision}{15 12}{}{11}.
\PrintBackRefs{\CurrentBib}

\bibitem [\protect \citeauthoryear {%
Brighton%
\ \BBA {} Kirby%
}{%
Brighton%
\ \BBA {} Kirby%
}{%
{\protect \APACyear {2006}}%
}]{%
Brighton2006UnderstandingLE}
\APACinsertmetastar {%
Brighton2006UnderstandingLE}%
\begin{APACrefauthors}%
Brighton, H.%
\BCBT {}\ \BBA {} Kirby, S.%
\end{APACrefauthors}%
\unskip\
\newblock
\APACrefYearMonthDay{2006}{}{}.
\newblock
{\BBOQ}\APACrefatitle {Understanding Linguistic Evolution by Visualizing the
  Emergence of Topographic Mappings} {Understanding linguistic evolution by
  visualizing the emergence of topographic mappings}.{\BBCQ}
\newblock
\APACjournalVolNumPages{Artificial Life}{12}{}{229-242}.
\PrintBackRefs{\CurrentBib}

\bibitem [\protect \citeauthoryear {%
Brinck%
}{%
Brinck%
}{%
{\protect \APACyear {2000}}%
}]{%
Brinck2000AttentionAT}
\APACinsertmetastar {%
Brinck2000AttentionAT}%
\begin{APACrefauthors}%
Brinck, I.%
\end{APACrefauthors}%
\unskip\
\newblock
\APACrefYearMonthDay{2000}{}{}.
\newblock
{\BBOQ}\APACrefatitle {Attention and the evolution of intentional
  communication} {Attention and the evolution of intentional
  communication}.{\BBCQ}
\newblock
\APACjournalVolNumPages{Pragmatics \& Cognition}{9}{}{255-272}.
\PrintBackRefs{\CurrentBib}

\bibitem [\protect \citeauthoryear {%
Chaabouni%
, Kharitonov%
, Bouchacourt%
, Dupoux%
\BCBL {}\ \BBA {} Baroni%
}{%
Chaabouni%
\ \protect \BOthers {.}}{%
{\protect \APACyear {2020}}%
}]{%
chaabouni-etal-2020-compositionality}
\APACinsertmetastar {%
chaabouni-etal-2020-compositionality}%
\begin{APACrefauthors}%
Chaabouni, R.%
, Kharitonov, E.%
, Bouchacourt, D.%
, Dupoux, E.%
\BCBL {}\ \BBA {} Baroni, M.%
\end{APACrefauthors}%
\unskip\
\newblock
\APACrefYearMonthDay{2020}{}{}.
\newblock
{\BBOQ}\APACrefatitle {Compositionality and Generalization In Emergent
  Languages} {Compositionality and generalization in emergent
  languages}.{\BBCQ}
\newblock
\BIn{} \APACrefbtitle {Proceedings of the 58th Annual Meeting of the
  Association for Computational Linguistics.} {Proceedings of the 58th annual
  meeting of the association for computational linguistics.}
\PrintBackRefs{\CurrentBib}

\bibitem [\protect \citeauthoryear {%
Chaabouni%
, Kharitonov%
, Lazaric%
, Dupoux%
\BCBL {}\ \BBA {} Baroni%
}{%
Chaabouni%
\ \protect \BOthers {.}}{%
{\protect \APACyear {2019}}%
}]{%
chaabouni-etal-2019-word}
\APACinsertmetastar {%
chaabouni-etal-2019-word}%
\begin{APACrefauthors}%
Chaabouni, R.%
, Kharitonov, E.%
, Lazaric, A.%
, Dupoux, E.%
\BCBL {}\ \BBA {} Baroni, M.%
\end{APACrefauthors}%
\unskip\
\newblock
\APACrefYearMonthDay{2019}{}{}.
\newblock
{\BBOQ}\APACrefatitle {Word-order Biases in Deep-agent Emergent Communication}
  {Word-order biases in deep-agent emergent communication}.{\BBCQ}
\newblock
\BIn{} \APACrefbtitle {Proceedings of the 57th Annual Meeting of the
  Association for Computational Linguistics.} {Proceedings of the 57th annual
  meeting of the association for computational linguistics.}
\PrintBackRefs{\CurrentBib}

\bibitem [\protect \citeauthoryear {%
Chaabouni%
\ \protect \BOthers {.}}{%
Chaabouni%
\ \protect \BOthers {.}}{%
{\protect \APACyear {2022}}%
}]{%
DBLP:conf/iclr/ChaabouniSATTDM22}
\APACinsertmetastar {%
DBLP:conf/iclr/ChaabouniSATTDM22}%
\begin{APACrefauthors}%
Chaabouni, R.%
, Strub, F.%
, Altch{\'{e}}, F.%
, Tarassov, E.%
, Tallec, C.%
, Davoodi, E.%
\BDBL {}Piot, B.%
\end{APACrefauthors}%
\unskip\
\newblock
\APACrefYearMonthDay{2022}{}{}.
\newblock
{\BBOQ}\APACrefatitle {Emergent Communication at Scale} {Emergent communication
  at scale}.{\BBCQ}
\newblock
\BIn{} \APACrefbtitle {International Conference on Learning Representations.}
  {International conference on learning representations.}
\PrintBackRefs{\CurrentBib}

\bibitem [\protect \citeauthoryear {%
de Diego-Balaguer%
, Martinez-Alvarez%
\BCBL {}\ \BBA {} Pons%
}{%
de Diego-Balaguer%
\ \protect \BOthers {.}}{%
{\protect \APACyear {2016}}%
}]{%
deDiegoBalaguer2016TemporalAA}
\APACinsertmetastar {%
deDiegoBalaguer2016TemporalAA}%
\begin{APACrefauthors}%
de Diego-Balaguer, R.%
, Martinez-Alvarez, A.%
\BCBL {}\ \BBA {} Pons, F.%
\end{APACrefauthors}%
\unskip\
\newblock
\APACrefYearMonthDay{2016}{}{}.
\newblock
{\BBOQ}\APACrefatitle {Temporal Attention as a Scaffold for Language
  Development} {Temporal attention as a scaffold for language
  development}.{\BBCQ}
\newblock
\APACjournalVolNumPages{Frontiers in Psychology}{7}{}{}.
\PrintBackRefs{\CurrentBib}

\bibitem [\protect \citeauthoryear {%
Estes%
, Verges%
\BCBL {}\ \BBA {} Barsalou%
}{%
Estes%
\ \protect \BOthers {.}}{%
{\protect \APACyear {2008}}%
}]{%
Estes2008HeadUF}
\APACinsertmetastar {%
Estes2008HeadUF}%
\begin{APACrefauthors}%
Estes, Z.%
, Verges, M.%
\BCBL {}\ \BBA {} Barsalou, L\BPBI W.%
\end{APACrefauthors}%
\unskip\
\newblock
\APACrefYearMonthDay{2008}{}{}.
\newblock
{\BBOQ}\APACrefatitle {Head Up, Foot Down} {Head up, foot down}.{\BBCQ}
\newblock
\APACjournalVolNumPages{Psychological Science}{19}{}{93 - 97}.
\PrintBackRefs{\CurrentBib}

\bibitem [\protect \citeauthoryear {%
Evtimova%
, Drozdov%
, Kiela%
\BCBL {}\ \BBA {} Cho%
}{%
Evtimova%
\ \protect \BOthers {.}}{%
{\protect \APACyear {2018}}%
}]{%
DBLP:conf/iclr/EvtimovaDKC18}
\APACinsertmetastar {%
DBLP:conf/iclr/EvtimovaDKC18}%
\begin{APACrefauthors}%
Evtimova, K.%
, Drozdov, A.%
, Kiela, D.%
\BCBL {}\ \BBA {} Cho, K.%
\end{APACrefauthors}%
\unskip\
\newblock
\APACrefYearMonthDay{2018}{}{}.
\newblock
{\BBOQ}\APACrefatitle {Emergent Communication in a Multi-Modal, Multi-Step
  Referential Game} {Emergent communication in a multi-modal, multi-step
  referential game}.{\BBCQ}
\newblock
\BIn{} \APACrefbtitle {International Conference on Learning Representations.}
  {International conference on learning representations.}
\PrintBackRefs{\CurrentBib}

\bibitem [\protect \citeauthoryear {%
Garrod%
\ \BBA {} Pickering%
}{%
Garrod%
\ \BBA {} Pickering%
}{%
{\protect \APACyear {2004}}%
}]{%
garrod_2004}
\APACinsertmetastar {%
garrod_2004}%
\begin{APACrefauthors}%
Garrod, S.%
\BCBT {}\ \BBA {} Pickering, M\BPBI J.%
\end{APACrefauthors}%
\unskip\
\newblock
\APACrefYearMonthDay{2004}{}{}.
\newblock
{\BBOQ}\APACrefatitle {Why is conversation so easy?} {Why is conversation so
  easy?}{\BBCQ}
\newblock
\APACjournalVolNumPages{Trends in Cognitive Sciences}{8}{1}{8 - 11}.
\newblock
\begin{APACrefDOI} \doi{DOI: 10.1016/j.tics.2003.10.016} \end{APACrefDOI}
\PrintBackRefs{\CurrentBib}

\bibitem [\protect \citeauthoryear {%
Hendrycks%
\ \BBA {} Gimpel%
}{%
Hendrycks%
\ \BBA {} Gimpel%
}{%
{\protect \APACyear {2016}}%
}]{%
Hendrycks2016GaussianEL}
\APACinsertmetastar {%
Hendrycks2016GaussianEL}%
\begin{APACrefauthors}%
Hendrycks, D.%
\BCBT {}\ \BBA {} Gimpel, K.%
\end{APACrefauthors}%
\unskip\
\newblock
\APACrefYearMonthDay{2016}{}{}.
\newblock
{\BBOQ}\APACrefatitle {Gaussian Error Linear Units (GELUs)} {Gaussian error
  linear units (gelus)}.{\BBCQ}
\newblock
\APACjournalVolNumPages{ArXiv}{abs/1606.08415}{}{}.
\PrintBackRefs{\CurrentBib}

\bibitem [\protect \citeauthoryear {%
Hochreiter%
\ \BBA {} Schmidhuber%
}{%
Hochreiter%
\ \BBA {} Schmidhuber%
}{%
{\protect \APACyear {1997}}%
}]{%
HochSchm97}
\APACinsertmetastar {%
HochSchm97}%
\begin{APACrefauthors}%
Hochreiter, S.%
\BCBT {}\ \BBA {} Schmidhuber, J.%
\end{APACrefauthors}%
\unskip\
\newblock
\APACrefYearMonthDay{1997}{}{}.
\newblock
{\BBOQ}\APACrefatitle {Long Short-Term Memory} {Long short-term memory}.{\BBCQ}
\newblock
\APACjournalVolNumPages{Neural Computation}{9}{8}{1735--1780}.
\PrintBackRefs{\CurrentBib}

\bibitem [\protect \citeauthoryear {%
Hudson%
\ \BBA {} Manning%
}{%
Hudson%
\ \BBA {} Manning%
}{%
{\protect \APACyear {2018}}%
}]{%
DBLP:conf/iclr/HudsonM18}
\APACinsertmetastar {%
DBLP:conf/iclr/HudsonM18}%
\begin{APACrefauthors}%
Hudson, D\BPBI A.%
\BCBT {}\ \BBA {} Manning, C\BPBI D.%
\end{APACrefauthors}%
\unskip\
\newblock
\APACrefYearMonthDay{2018}{}{}.
\newblock
{\BBOQ}\APACrefatitle {Compositional Attention Networks for Machine Reasoning}
  {Compositional attention networks for machine reasoning}.{\BBCQ}
\newblock
\BIn{} \APACrefbtitle {International Conference on Learning Representations.}
  {International conference on learning representations.}
\PrintBackRefs{\CurrentBib}

\bibitem [\protect \citeauthoryear {%
Hudson%
\ \BBA {} Zitnick%
}{%
Hudson%
\ \BBA {} Zitnick%
}{%
{\protect \APACyear {2021}}%
}]{%
Hudson2021CompositionalTF}
\APACinsertmetastar {%
Hudson2021CompositionalTF}%
\begin{APACrefauthors}%
Hudson, D\BPBI A.%
\BCBT {}\ \BBA {} Zitnick, C\BPBI L.%
\end{APACrefauthors}%
\unskip\
\newblock
\APACrefYearMonthDay{2021}{}{}.
\newblock
{\BBOQ}\APACrefatitle {Compositional Transformers for Scene Generation}
  {Compositional transformers for scene generation}.{\BBCQ}
\newblock
\BIn{} \APACrefbtitle {Advances in Neural Information Processing Systems.}
  {Advances in neural information processing systems.}
\PrintBackRefs{\CurrentBib}

\bibitem [\protect \citeauthoryear {%
Jain%
\ \BBA {} Wallace%
}{%
Jain%
\ \BBA {} Wallace%
}{%
{\protect \APACyear {2019}}%
}]{%
jain-wallace-2019-attention}
\APACinsertmetastar {%
jain-wallace-2019-attention}%
\begin{APACrefauthors}%
Jain, S.%
\BCBT {}\ \BBA {} Wallace, B\BPBI C.%
\end{APACrefauthors}%
\unskip\
\newblock
\APACrefYearMonthDay{2019}{}{}.
\newblock
{\BBOQ}\APACrefatitle {{A}ttention is not {E}xplanation} {{A}ttention is not
  {E}xplanation}.{\BBCQ}
\newblock
\BIn{} \APACrefbtitle {Proceedings of the 2019 Conference of the North
  {A}merican Chapter of the Association for Computational Linguistics: Human
  Language Technologies, Volume 1 (Long and Short Papers).} {Proceedings of the
  2019 conference of the north {A}merican chapter of the association for
  computational linguistics: Human language technologies, volume 1 (long and
  short papers).}
\PrintBackRefs{\CurrentBib}

\bibitem [\protect \citeauthoryear {%
Korrel%
, Hupkes%
, Dankers%
\BCBL {}\ \BBA {} Bruni%
}{%
Korrel%
\ \protect \BOthers {.}}{%
{\protect \APACyear {2019}}%
}]{%
korrel-etal-2019-transcoding}
\APACinsertmetastar {%
korrel-etal-2019-transcoding}%
\begin{APACrefauthors}%
Korrel, K.%
, Hupkes, D.%
, Dankers, V.%
\BCBL {}\ \BBA {} Bruni, E.%
\end{APACrefauthors}%
\unskip\
\newblock
\APACrefYearMonthDay{2019}{}{}.
\newblock
{\BBOQ}\APACrefatitle {Transcoding Compositionally: Using Attention to Find
  More Generalizable Solutions} {Transcoding compositionally: Using attention
  to find more generalizable solutions}.{\BBCQ}
\newblock
\BIn{} \APACrefbtitle {Proceedings of the 2019 ACL Workshop {BlackboxNLP}:
  Analyzing and Interpreting Neural Networks for {NLP}.} {Proceedings of the
  2019 acl workshop {BlackboxNLP}: Analyzing and interpreting neural networks
  for {NLP}.}
\PrintBackRefs{\CurrentBib}

\bibitem [\protect \citeauthoryear {%
Kriegeskorte%
, Mur%
\BCBL {}\ \BBA {} Bandettini%
}{%
Kriegeskorte%
\ \protect \BOthers {.}}{%
{\protect \APACyear {2008}}%
}]{%
Kriegeskorte2008}
\APACinsertmetastar {%
Kriegeskorte2008}%
\begin{APACrefauthors}%
Kriegeskorte, N.%
, Mur, M.%
\BCBL {}\ \BBA {} Bandettini, P.%
\end{APACrefauthors}%
\unskip\
\newblock
\APACrefYearMonthDay{2008}{}{}.
\newblock
{\BBOQ}\APACrefatitle {Representational similarity analysis - connecting the
  branches of systems neuroscience.} {Representational similarity analysis -
  connecting the branches of systems neuroscience.}{\BBCQ}
\newblock
\APACjournalVolNumPages{Frontiers in systems neuroscience}{2}{}{}.
\PrintBackRefs{\CurrentBib}

\bibitem [\protect \citeauthoryear {%
Kwisthout%
, Vogt%
, Haselager%
\BCBL {}\ \BBA {} Dijkstra%
}{%
Kwisthout%
\ \protect \BOthers {.}}{%
{\protect \APACyear {2008}}%
}]{%
Kwisthout2008JointAA}
\APACinsertmetastar {%
Kwisthout2008JointAA}%
\begin{APACrefauthors}%
Kwisthout, J.%
, Vogt, P.%
, Haselager, W\BPBI P.%
\BCBL {}\ \BBA {} Dijkstra, T.%
\end{APACrefauthors}%
\unskip\
\newblock
\APACrefYearMonthDay{2008}{}{}.
\newblock
{\BBOQ}\APACrefatitle {Joint attention and language evolution} {Joint attention
  and language evolution}.{\BBCQ}
\newblock
\APACjournalVolNumPages{Connection Science}{20}{}{155 - 171}.
\PrintBackRefs{\CurrentBib}

\bibitem [\protect \citeauthoryear {%
Lazaridou%
\ \BBA {} Baroni%
}{%
Lazaridou%
\ \BBA {} Baroni%
}{%
{\protect \APACyear {2020}}%
}]{%
Lazaridou2020EmergentMC}
\APACinsertmetastar {%
Lazaridou2020EmergentMC}%
\begin{APACrefauthors}%
Lazaridou, A.%
\BCBT {}\ \BBA {} Baroni, M.%
\end{APACrefauthors}%
\unskip\
\newblock
\APACrefYearMonthDay{2020}{}{}.
\newblock
{\BBOQ}\APACrefatitle {Emergent Multi-Agent Communication in the Deep Learning
  Era} {Emergent multi-agent communication in the deep learning era}.{\BBCQ}
\newblock
\APACjournalVolNumPages{ArXiv}{abs/2006.02419}{}{}.
\PrintBackRefs{\CurrentBib}

\bibitem [\protect \citeauthoryear {%
Lazaridou%
, Hermann%
, Tuyls%
\BCBL {}\ \BBA {} Clark%
}{%
Lazaridou%
\ \protect \BOthers {.}}{%
{\protect \APACyear {2018}}%
}]{%
DBLP:conf/iclr/LazaridouHTC18}
\APACinsertmetastar {%
DBLP:conf/iclr/LazaridouHTC18}%
\begin{APACrefauthors}%
Lazaridou, A.%
, Hermann, K\BPBI M.%
, Tuyls, K.%
\BCBL {}\ \BBA {} Clark, S.%
\end{APACrefauthors}%
\unskip\
\newblock
\APACrefYearMonthDay{2018}{}{}.
\newblock
{\BBOQ}\APACrefatitle {Emergence of Linguistic Communication from Referential
  Games with Symbolic and Pixel Input} {Emergence of linguistic communication
  from referential games with symbolic and pixel input}.{\BBCQ}
\newblock
\BIn{} \APACrefbtitle {International Conference on Learning Representations.}
  {International conference on learning representations.}
\PrintBackRefs{\CurrentBib}

\bibitem [\protect \citeauthoryear {%
Lazaridou%
, Peysakhovich%
\BCBL {}\ \BBA {} Baroni%
}{%
Lazaridou%
\ \protect \BOthers {.}}{%
{\protect \APACyear {2017}}%
}]{%
DBLP:conf/iclr/LazaridouPB17}
\APACinsertmetastar {%
DBLP:conf/iclr/LazaridouPB17}%
\begin{APACrefauthors}%
Lazaridou, A.%
, Peysakhovich, A.%
\BCBL {}\ \BBA {} Baroni, M.%
\end{APACrefauthors}%
\unskip\
\newblock
\APACrefYearMonthDay{2017}{}{}.
\newblock
{\BBOQ}\APACrefatitle {Multi-Agent Cooperation and the Emergence of (Natural)
  Language} {Multi-agent cooperation and the emergence of (natural)
  language}.{\BBCQ}
\newblock
\BIn{} \APACrefbtitle {International Conference on Learning Representations.}
  {International conference on learning representations.}
\PrintBackRefs{\CurrentBib}

\bibitem [\protect \citeauthoryear {%
Lewis%
}{%
Lewis%
}{%
{\protect \APACyear {1969}}%
}]{%
Lewis1969-LEWCAP-4}
\APACinsertmetastar {%
Lewis1969-LEWCAP-4}%
\begin{APACrefauthors}%
Lewis, D\BPBI K.%
\end{APACrefauthors}%
\unskip\
\newblock
\APACrefYear{1969}.
\newblock
\APACrefbtitle {Convention: A Philosophical Study} {Convention: A philosophical
  study}.
\newblock
\APACaddressPublisher{}{Cambridge, MA, USA: Wiley-Blackwell}.
\PrintBackRefs{\CurrentBib}

\bibitem [\protect \citeauthoryear {%
Li%
\ \BBA {} Bowling%
}{%
Li%
\ \BBA {} Bowling%
}{%
{\protect \APACyear {2019}}%
}]{%
NEURIPS2019_b0cf188d}
\APACinsertmetastar {%
NEURIPS2019_b0cf188d}%
\begin{APACrefauthors}%
Li, F.%
\BCBT {}\ \BBA {} Bowling, M.%
\end{APACrefauthors}%
\unskip\
\newblock
\APACrefYearMonthDay{2019}{}{}.
\newblock
{\BBOQ}\APACrefatitle {Ease-of-Teaching and Language Structure from Emergent
  Communication} {Ease-of-teaching and language structure from emergent
  communication}.{\BBCQ}
\newblock
\BIn{} \APACrefbtitle {Advances in Neural Information Processing Systems}
  {Advances in neural information processing systems}\ (\BVOL~32).
\PrintBackRefs{\CurrentBib}

\bibitem [\protect \citeauthoryear {%
Lin%
}{%
Lin%
}{%
{\protect \APACyear {1991}}%
}]{%
Lin1991DivergenceMB}
\APACinsertmetastar {%
Lin1991DivergenceMB}%
\begin{APACrefauthors}%
Lin, J.%
\end{APACrefauthors}%
\unskip\
\newblock
\APACrefYearMonthDay{1991}{}{}.
\newblock
{\BBOQ}\APACrefatitle {Divergence measures based on the Shannon entropy}
  {Divergence measures based on the shannon entropy}.{\BBCQ}
\newblock
\APACjournalVolNumPages{{IEEE} Transactions on Information
  Theory}{37}{}{145-151}.
\PrintBackRefs{\CurrentBib}

\bibitem [\protect \citeauthoryear {%
F.~Liu%
\ \protect \BOthers {.}}{%
F.~Liu%
\ \protect \BOthers {.}}{%
{\protect \APACyear {2020}}%
}]{%
Liu2020ProphetAP}
\APACinsertmetastar {%
Liu2020ProphetAP}%
\begin{APACrefauthors}%
Liu, F.%
, Ren, X.%
, Wu, X.%
, Ge, S.%
, Fan, W.%
, Zou, Y.%
\BCBL {}\ \BBA {} Sun, X.%
\end{APACrefauthors}%
\unskip\
\newblock
\APACrefYearMonthDay{2020}{}{}.
\newblock
{\BBOQ}\APACrefatitle {Prophet Attention: Predicting Attention with Future
  Attention} {Prophet attention: Predicting attention with future
  attention}.{\BBCQ}
\newblock
\BIn{} \APACrefbtitle {Advances in Neural Information Processing Systems.}
  {Advances in neural information processing systems.}
\PrintBackRefs{\CurrentBib}

\bibitem [\protect \citeauthoryear {%
Z.~Liu%
\ \protect \BOthers {.}}{%
Z.~Liu%
\ \protect \BOthers {.}}{%
{\protect \APACyear {2022}}%
}]{%
liu2022convnet}
\APACinsertmetastar {%
liu2022convnet}%
\begin{APACrefauthors}%
Liu, Z.%
, Mao, H.%
, Wu, C\BHBI Y.%
, Feichtenhofer, C.%
, Darrell, T.%
\BCBL {}\ \BBA {} Xie, S.%
\end{APACrefauthors}%
\unskip\
\newblock
\APACrefYearMonthDay{2022}{}{}.
\newblock
{\BBOQ}\APACrefatitle {A ConvNet for the 2020s} {A convnet for the
  2020s}.{\BBCQ}
\newblock
\APACjournalVolNumPages{Proceedings of the {IEEE/CVF} Conference on Computer
  Vision and Pattern Recognition ({CVPR})}{}{}{}.
\PrintBackRefs{\CurrentBib}

\bibitem [\protect \citeauthoryear {%
Lowe%
, Foerster%
, Boureau%
, Pineau%
\BCBL {}\ \BBA {} Dauphin%
}{%
Lowe%
\ \protect \BOthers {.}}{%
{\protect \APACyear {2019}}%
}]{%
Lowe2019OnTP}
\APACinsertmetastar {%
Lowe2019OnTP}%
\begin{APACrefauthors}%
Lowe, R.%
, Foerster, J\BPBI N.%
, Boureau, Y\BHBI L.%
, Pineau, J.%
\BCBL {}\ \BBA {} Dauphin, Y.%
\end{APACrefauthors}%
\unskip\
\newblock
\APACrefYearMonthDay{2019}{}{}.
\newblock
{\BBOQ}\APACrefatitle {On the Pitfalls of Measuring Emergent Communication} {On
  the pitfalls of measuring emergent communication}.{\BBCQ}
\newblock
\BIn{} \APACrefbtitle {AAMAS.} {Aamas.}
\PrintBackRefs{\CurrentBib}

\bibitem [\protect \citeauthoryear {%
Luong%
, Pham%
\BCBL {}\ \BBA {} Manning%
}{%
Luong%
\ \protect \BOthers {.}}{%
{\protect \APACyear {2015}}%
}]{%
luong-etal-2015-effective}
\APACinsertmetastar {%
luong-etal-2015-effective}%
\begin{APACrefauthors}%
Luong, T.%
, Pham, H.%
\BCBL {}\ \BBA {} Manning, C\BPBI D.%
\end{APACrefauthors}%
\unskip\
\newblock
\APACrefYearMonthDay{2015}{}{}.
\newblock
{\BBOQ}\APACrefatitle {Effective Approaches to Attention-based Neural Machine
  Translation} {Effective approaches to attention-based neural machine
  translation}.{\BBCQ}
\newblock
\BIn{} \APACrefbtitle {Proceedings of the 2015 Conference on Empirical Methods
  in Natural Language Processing.} {Proceedings of the 2015 conference on
  empirical methods in natural language processing.}
\PrintBackRefs{\CurrentBib}

\bibitem [\protect \citeauthoryear {%
Mordatch%
\ \BBA {} Abbeel%
}{%
Mordatch%
\ \BBA {} Abbeel%
}{%
{\protect \APACyear {2018}}%
}]{%
DBLP:conf/aaai/MordatchA18}
\APACinsertmetastar {%
DBLP:conf/aaai/MordatchA18}%
\begin{APACrefauthors}%
Mordatch, I.%
\BCBT {}\ \BBA {} Abbeel, P.%
\end{APACrefauthors}%
\unskip\
\newblock
\APACrefYearMonthDay{2018}{}{}.
\newblock
{\BBOQ}\APACrefatitle {Emergence of Grounded Compositional Language in
  Multi-Agent Populations} {Emergence of grounded compositional language in
  multi-agent populations}.{\BBCQ}
\newblock
\BIn{} \APACrefbtitle {The Thirty-Second AAAI Conference on Artificial
  Intelligence.} {The thirty-second aaai conference on artificial
  intelligence.}
\PrintBackRefs{\CurrentBib}

\bibitem [\protect \citeauthoryear {%
Mu%
\ \BBA {} Goodman%
}{%
Mu%
\ \BBA {} Goodman%
}{%
{\protect \APACyear {2021}}%
}]{%
DBLP:conf/nips/MuG21}
\APACinsertmetastar {%
DBLP:conf/nips/MuG21}%
\begin{APACrefauthors}%
Mu, J.%
\BCBT {}\ \BBA {} Goodman, N\BPBI D.%
\end{APACrefauthors}%
\unskip\
\newblock
\APACrefYearMonthDay{2021}{}{}.
\newblock
{\BBOQ}\APACrefatitle {Emergent Communication of Generalizations} {Emergent
  communication of generalizations}.{\BBCQ}
\newblock
\BIn{} \APACrefbtitle {Advances in Neural Information Processing Systems}
  {Advances in neural information processing systems}\ (\BPGS\ 17994--18007).
\PrintBackRefs{\CurrentBib}

\bibitem [\protect \citeauthoryear {%
Ren%
, Guo%
, Labeau%
, Cohen%
\BCBL {}\ \BBA {} Kirby%
}{%
Ren%
\ \protect \BOthers {.}}{%
{\protect \APACyear {2020}}%
}]{%
DBLP:conf/iclr/RenGLCK20}
\APACinsertmetastar {%
DBLP:conf/iclr/RenGLCK20}%
\begin{APACrefauthors}%
Ren, Y.%
, Guo, S.%
, Labeau, M.%
, Cohen, S\BPBI B.%
\BCBL {}\ \BBA {} Kirby, S.%
\end{APACrefauthors}%
\unskip\
\newblock
\APACrefYearMonthDay{2020}{}{}.
\newblock
{\BBOQ}\APACrefatitle {Compositional languages emerge in a neural iterated
  learning model} {Compositional languages emerge in a neural iterated learning
  model}.{\BBCQ}
\newblock
\BIn{} \APACrefbtitle {International Conference on Learning Representations.}
  {International conference on learning representations.}
\PrintBackRefs{\CurrentBib}

\bibitem [\protect \citeauthoryear {%
Rensink%
}{%
Rensink%
}{%
{\protect \APACyear {2000}}%
}]{%
Rensink2000TheDR}
\APACinsertmetastar {%
Rensink2000TheDR}%
\begin{APACrefauthors}%
Rensink, R\BPBI A.%
\end{APACrefauthors}%
\unskip\
\newblock
\APACrefYearMonthDay{2000}{}{}.
\newblock
{\BBOQ}\APACrefatitle {The Dynamic Representation of Scenes} {The dynamic
  representation of scenes}.{\BBCQ}
\newblock
\APACjournalVolNumPages{Visual Cognition}{7}{}{17 - 42}.
\PrintBackRefs{\CurrentBib}

\bibitem [\protect \citeauthoryear {%
Rita%
, Strub%
, Grill%
, Pietquin%
\BCBL {}\ \BBA {} Dupoux%
}{%
Rita%
, Strub%
\BCBL {}\ \protect \BOthers {.}}{%
{\protect \APACyear {2022}}%
}]{%
DBLP:conf/iclr/RitaSGPD22}
\APACinsertmetastar {%
DBLP:conf/iclr/RitaSGPD22}%
\begin{APACrefauthors}%
Rita, M.%
, Strub, F.%
, Grill, J.%
, Pietquin, O.%
\BCBL {}\ \BBA {} Dupoux, E.%
\end{APACrefauthors}%
\unskip\
\newblock
\APACrefYearMonthDay{2022}{}{}.
\newblock
{\BBOQ}\APACrefatitle {On the role of population heterogeneity in emergent
  communication} {On the role of population heterogeneity in emergent
  communication}.{\BBCQ}
\newblock
\BIn{} \APACrefbtitle {International Conference on Learning Representations.}
  {International conference on learning representations.}
\PrintBackRefs{\CurrentBib}

\bibitem [\protect \citeauthoryear {%
Rita%
, Tallec%
\BCBL {}\ \protect \BOthers {.}}{%
Rita%
, Tallec%
\BCBL {}\ \protect \BOthers {.}}{%
{\protect \APACyear {2022}}%
}]{%
Rita2022EmergentCG}
\APACinsertmetastar {%
Rita2022EmergentCG}%
\begin{APACrefauthors}%
Rita, M.%
, Tallec, C.%
, Michel, P.%
, Grill, J\BHBI B.%
, Pietquin, O.%
, Dupoux, E.%
\BCBL {}\ \BBA {} Strub, F.%
\end{APACrefauthors}%
\unskip\
\newblock
\APACrefYearMonthDay{2022}{}{}.
\newblock
{\BBOQ}\APACrefatitle {Emergent Communication: Generalization and Overfitting
  in Lewis Games} {Emergent communication: Generalization and overfitting in
  lewis games}.{\BBCQ}
\newblock
\BIn{} \APACrefbtitle {Advances in Neural Information Processing Systems.}
  {Advances in neural information processing systems.}
\PrintBackRefs{\CurrentBib}

\bibitem [\protect \citeauthoryear {%
Rodr{\'\i}guez~Luna%
, Ponti%
, Hupkes%
\BCBL {}\ \BBA {} Bruni%
}{%
Rodr{\'\i}guez~Luna%
\ \protect \BOthers {.}}{%
{\protect \APACyear {2020}}%
}]{%
rodriguez-luna-etal-2020-internal}
\APACinsertmetastar {%
rodriguez-luna-etal-2020-internal}%
\begin{APACrefauthors}%
Rodr{\'\i}guez~Luna, D.%
, Ponti, E\BPBI M.%
, Hupkes, D.%
\BCBL {}\ \BBA {} Bruni, E.%
\end{APACrefauthors}%
\unskip\
\newblock
\APACrefYearMonthDay{2020}{}{}.
\newblock
{\BBOQ}\APACrefatitle {Internal and external pressures on language emergence:
  least effort, object constancy and frequency} {Internal and external
  pressures on language emergence: least effort, object constancy and
  frequency}.{\BBCQ}
\newblock
\BIn{} \APACrefbtitle {Findings of the Association for Computational
  Linguistics: {EMNLP} 2020.} {Findings of the association for computational
  linguistics: {EMNLP} 2020.}
\PrintBackRefs{\CurrentBib}

\bibitem [\protect \citeauthoryear {%
Ryo%
, Taiga%
, Koki%
\BCBL {}\ \BBA {} Yusuke%
}{%
Ryo%
\ \protect \BOthers {.}}{%
{\protect \APACyear {2022}}%
}]{%
ueda-etal-2022-cgi}
\APACinsertmetastar {%
ueda-etal-2022-cgi}%
\begin{APACrefauthors}%
Ryo, U.%
, Taiga, I.%
, Koki, W.%
\BCBL {}\ \BBA {} Yusuke, M.%
\end{APACrefauthors}%
\unskip\
\newblock
\APACrefYearMonthDay{2022}{}{}.
\newblock
{\BBOQ}\APACrefatitle {Categorial Grammar Induction as a Compositionality
  Measure for Emergent Languages in Signaling Games} {Categorial grammar
  induction as a compositionality measure for emergent languages in signaling
  games}.{\BBCQ}
\newblock
\BIn{} \APACrefbtitle {Proceedings of Emergent Communication Workshop at ICLR
  2022.} {Proceedings of emergent communication workshop at iclr 2022.}
\PrintBackRefs{\CurrentBib}

\bibitem [\protect \citeauthoryear {%
Słowik%
\ \protect \BOthers {.}}{%
Słowik%
\ \protect \BOthers {.}}{%
{\protect \APACyear {2020}}%
}]{%
Sowik2020StructuralIB}
\APACinsertmetastar {%
Sowik2020StructuralIB}%
\begin{APACrefauthors}%
Słowik, A.%
, Gupta, A\BPBI K.%
, Hamilton, W\BPBI L.%
, Jamnik, M.%
, Holden, S.%
\BCBL {}\ \BBA {} Pal, C\BPBI J.%
\end{APACrefauthors}%
\unskip\
\newblock
\APACrefYearMonthDay{2020}{}{}.
\newblock
{\BBOQ}\APACrefatitle {Structural Inductive Biases in Emergent Communication}
  {Structural inductive biases in emergent communication}.{\BBCQ}
\newblock
\BIn{} \APACrefbtitle {Proceedings of The 43rd Annual Meeting of the Cognitive
  Science Society.} {Proceedings of the 43rd annual meeting of the cognitive
  science society.}
\PrintBackRefs{\CurrentBib}

\bibitem [\protect \citeauthoryear {%
Vaswani%
\ \protect \BOthers {.}}{%
Vaswani%
\ \protect \BOthers {.}}{%
{\protect \APACyear {2017}}%
}]{%
DBLP:conf/nips/VaswaniSPUJGKP17}
\APACinsertmetastar {%
DBLP:conf/nips/VaswaniSPUJGKP17}%
\begin{APACrefauthors}%
Vaswani, A.%
, Shazeer, N.%
, Parmar, N.%
, Uszkoreit, J.%
, Jones, L.%
, Gomez, A\BPBI N.%
\BDBL {}Polosukhin, I.%
\end{APACrefauthors}%
\unskip\
\newblock
\APACrefYearMonthDay{2017}{}{}.
\newblock
{\BBOQ}\APACrefatitle {Attention is All you Need} {Attention is all you
  need}.{\BBCQ}
\newblock
\BIn{} \APACrefbtitle {Advances in Neural Information Processing Systems}
  {Advances in neural information processing systems}\ (\BVOL~30).
\PrintBackRefs{\CurrentBib}

\bibitem [\protect \citeauthoryear {%
Wiegreffe%
\ \BBA {} Pinter%
}{%
Wiegreffe%
\ \BBA {} Pinter%
}{%
{\protect \APACyear {2019}}%
}]{%
wiegreffe-pinter-2019-attention}
\APACinsertmetastar {%
wiegreffe-pinter-2019-attention}%
\begin{APACrefauthors}%
Wiegreffe, S.%
\BCBT {}\ \BBA {} Pinter, Y.%
\end{APACrefauthors}%
\unskip\
\newblock
\APACrefYearMonthDay{2019}{}{}.
\newblock
{\BBOQ}\APACrefatitle {Attention is not not Explanation} {Attention is not not
  explanation}.{\BBCQ}
\newblock
\BIn{} \APACrefbtitle {Proceedings of the 2019 Conference on Empirical Methods
  in Natural Language Processing and the 9th International Joint Conference on
  Natural Language Processing ({EMNLP-IJCNLP)}.} {Proceedings of the 2019
  conference on empirical methods in natural language processing and the 9th
  international joint conference on natural language processing
  ({EMNLP-IJCNLP)}.}
\PrintBackRefs{\CurrentBib}

\bibitem [\protect \citeauthoryear {%
Williams%
}{%
Williams%
}{%
{\protect \APACyear {1992}}%
}]{%
REINFORCE1992}
\APACinsertmetastar {%
REINFORCE1992}%
\begin{APACrefauthors}%
Williams, R\BPBI J.%
\end{APACrefauthors}%
\unskip\
\newblock
\APACrefYearMonthDay{1992}{}{}.
\newblock
{\BBOQ}\APACrefatitle {Simple Statistical Gradient-Following Algorithms for
  Connectionist Reinforcement Learning} {Simple statistical gradient-following
  algorithms for connectionist reinforcement learning}.{\BBCQ}
\newblock
\APACjournalVolNumPages{Machine Learning}{8}{3–4}{229–256}.
\PrintBackRefs{\CurrentBib}

\bibitem [\protect \citeauthoryear {%
Xiao%
, Rasul%
\BCBL {}\ \BBA {} Vollgraf%
}{%
Xiao%
\ \protect \BOthers {.}}{%
{\protect \APACyear {2017}}%
}]{%
Xiao2017FashionMNISTAN}
\APACinsertmetastar {%
Xiao2017FashionMNISTAN}%
\begin{APACrefauthors}%
Xiao, H.%
, Rasul, K.%
\BCBL {}\ \BBA {} Vollgraf, R.%
\end{APACrefauthors}%
\unskip\
\newblock
\APACrefYearMonthDay{2017}{}{}.
\newblock
{\BBOQ}\APACrefatitle {Fashion-MNIST: a Novel Image Dataset for Benchmarking
  Machine Learning Algorithms} {Fashion-mnist: a novel image dataset for
  benchmarking machine learning algorithms}.{\BBCQ}
\newblock
\APACjournalVolNumPages{ArXiv}{abs/1708.07747}{}{}.
\PrintBackRefs{\CurrentBib}

\bibitem [\protect \citeauthoryear {%
Xu%
\ \protect \BOthers {.}}{%
Xu%
\ \protect \BOthers {.}}{%
{\protect \APACyear {2015}}%
}]{%
DBLP:conf/icml/XuBKCCSZB15}
\APACinsertmetastar {%
DBLP:conf/icml/XuBKCCSZB15}%
\begin{APACrefauthors}%
Xu, K.%
, Ba, J.%
, Kiros, R.%
, Cho, K.%
, Courville, A\BPBI C.%
, Salakhutdinov, R.%
\BDBL {}Bengio, Y.%
\end{APACrefauthors}%
\unskip\
\newblock
\APACrefYearMonthDay{2015}{}{}.
\newblock
{\BBOQ}\APACrefatitle {Show, Attend and Tell: Neural Image Caption Generation
  with Visual Attention} {Show, attend and tell: Neural image caption
  generation with visual attention}.{\BBCQ}
\newblock
\BIn{} \APACrefbtitle {Proceedings of the International Conference on Machine
  Learning} {Proceedings of the international conference on machine learning}\
  (\BVOL~37).
\PrintBackRefs{\CurrentBib}

\bibitem [\protect \citeauthoryear {%
Yang%
, He%
, Gao%
, Deng%
\BCBL {}\ \BBA {} Smola%
}{%
Yang%
\ \protect \BOthers {.}}{%
{\protect \APACyear {2016}}%
}]{%
Yang2016StackedAN}
\APACinsertmetastar {%
Yang2016StackedAN}%
\begin{APACrefauthors}%
Yang, Z.%
, He, X.%
, Gao, J.%
, Deng, L.%
\BCBL {}\ \BBA {} Smola, A.%
\end{APACrefauthors}%
\unskip\
\newblock
\APACrefYearMonthDay{2016}{}{}.
\newblock
{\BBOQ}\APACrefatitle {Stacked Attention Networks for Image Question Answering}
  {Stacked attention networks for image question answering}.{\BBCQ}
\newblock
\BIn{} \APACrefbtitle {{IEEE} Conference on Computer Vision and Pattern
  Recognition ({CVPR}).} {{IEEE} conference on computer vision and pattern
  recognition ({CVPR}).}
\PrintBackRefs{\CurrentBib}

\bibitem [\protect \citeauthoryear {%
Łukasz Kuciński%
, Korbak%
, Kołodziej%
\BCBL {}\ \BBA {} Miłoś%
}{%
Łukasz Kuciński%
\ \protect \BOthers {.}}{%
{\protect \APACyear {2021}}%
}]{%
NEURIPS2021_c2839bed}
\APACinsertmetastar {%
NEURIPS2021_c2839bed}%
\begin{APACrefauthors}%
Łukasz Kuciński%
, Korbak, T.%
, Kołodziej, P.%
\BCBL {}\ \BBA {} Miłoś, P.%
\end{APACrefauthors}%
\unskip\
\newblock
\APACrefYearMonthDay{2021}{}{}.
\newblock
{\BBOQ}\APACrefatitle {Catalytic Role Of Noise And Necessity Of Inductive
  Biases In The Emergence Of Compositional Communication} {Catalytic role of
  noise and necessity of inductive biases in the emergence of compositional
  communication}.{\BBCQ}
\newblock
\BIn{} \APACrefbtitle {Advances in Neural Information Processing Systems}
  {Advances in neural information processing systems}\ (\BVOL~34).
\PrintBackRefs{\CurrentBib}

\end{thebibliography}

\end{document}